\pdfoutput=1

\documentclass[11pt]{article}
\usepackage{graphicx}
\usepackage{ACL2023}
\usepackage{forest}
\usetikzlibrary{shadows}
\definecolorseries{colours}{hsb}{grad}[hsb]{.575,1,1}{.987,-.234,0}
\resetcolorseries[12]{colours}
\usepackage{times}
\usepackage{amsmath}
\usepackage{latexsym}
 \usepackage{multirow}
 \usepackage{graphicx}
\usepackage{soul}
\usepackage[T1]{fontenc}
\usepackage{lipsum}

\usepackage[utf8]{inputenc}
\usepackage{microtype}

\usepackage{inconsolata}

\usepackage{lipsum} 
\usepackage{xcolor}
\definecolor{darkgreen}{rgb}{0.0, 0.5, 0.0}

\title{Automated Justification Production for Claim Veracity in Fact Checking: A Survey on Architectures and Approaches}

\author{
  \textbf{Islam Eldifrawi},
  ~\textbf{Shengrui Wang},
  ~\textbf{Amine Trabelsi} \\
Department of Computer Science,
Université de Sherbrooke \\
    \normalsize{\tt{\{Islam.Eldifrawi;Shengrui.Wang;Amine.Trabelsi\}@usherbrooke.ca} }
}

\begin{document}
\graphicspath{ {./images/} }
\maketitle
\begin{abstract}

Automated Fact-Checking (AFC) is the automated verification of claim accuracy. AFC is crucial in discerning truth from misinformation, especially given the huge amounts of content are generated online daily. Current research focuses on predicting claim veracity through metadata analysis and language scrutiny, with an emphasis on justifying verdicts. This paper surveys recent methodologies, proposing a comprehensive taxonomy and presenting the evolution of research in that landscape. A comparative analysis of methodologies and future directions for improving fact-checking explainability are also discussed. 

\end{abstract}

\section{Introduction}

The huge increase in both user-generated and automated content has led to a significant amount of misinformation. This poses risks to uninformed readers, highlighting the need for scalable, automated methods for verification and fact-checking  \cite{nakov2021automated}. While predicting the veracity of claims is essential, relying solely on predictions without providing explanations can be counterproductive, potentially reinforcing belief in false claims and perpetuating misinformation \cite{lewandowsky2012misinformation}. 

Most fact-checking models use neural architectures, but interpreting these models is challenging. There is a need for fact-checking frameworks providing justifications to enhance \textbf{effectiveness} and \textbf{trustworthiness}. This survey presents recent efforts addressing automatic justification production for claim verification, emphasizing the move towards ``\textbf{Explainable}'' \textbf{Automated Fact-Checking} (\textbf{AFC}). 
Some work refers to the justification production process as the explanation generation process \cite{kotonya-toni-2020-explainable}. In this survey, the term ``justification production'' is used following the work of \citet{guo-etal-2022-survey}. 

This survey's main contribution is as follows: Firstly, it introduces a multidimensional taxonomy for categorizing works based on various criteria. Secondly, it provides how research is progressing towards standard justifications. Thirdly, it conducts a comparative analysis of justification production approaches, pipeline architectures, input and output types. Lastly, it identifies challenges while proposing future directions in justification production.
Appendix \ref{const} outlines the methodology utilized for literature compilation, detailing the search strategy and selection criteria employed for the papers that form the cornerstone of this survey.

 
\section{Related Surveys}
\label{r_sur}
\citet{thorne-vlachos-2018-automated} provided a comprehensive review of early developments in fact-checking, but they don't focus on verdicts with \textbf{justifications}. Other surveys such as \citet{ijcai2021p619, nakov2021automated, guo-etal-2022-survey, vladika-matthes-2023-scientific} offer broad overviews of the entire fact-checking process and its various components. In contrast, our work specifically concentrates on the aspect of justification production. Moreover, recent multi-modal fact-checking surveys \cite{alam-etal-2022-survey, vladika-matthes-2023-scientific} mention that natural language justification production remains unexplored in the multi-modal AFC domain. In this survey, we highlight some of the emergent works in multi-modal justification production.

The survey by \citet{kotonya-toni-2020-explainable}, focusing on justification production, is closely related to ours. However, since then, there has been a significant progress driven by the rapid development of transformer-based architectures and Large Language Models (LLMs). \citet{vallayil2023explainability} only augmented the latter work's taxonomy with counterfactual justifications. While partially covering some recent work, they do not provide a comprehensive, new, detailed multi-dimensional taxonomy as proposed in this survey.

\section{Justification Production within AFC}
\begin{figure*}
    \centering
    \includegraphics[width=10cm]{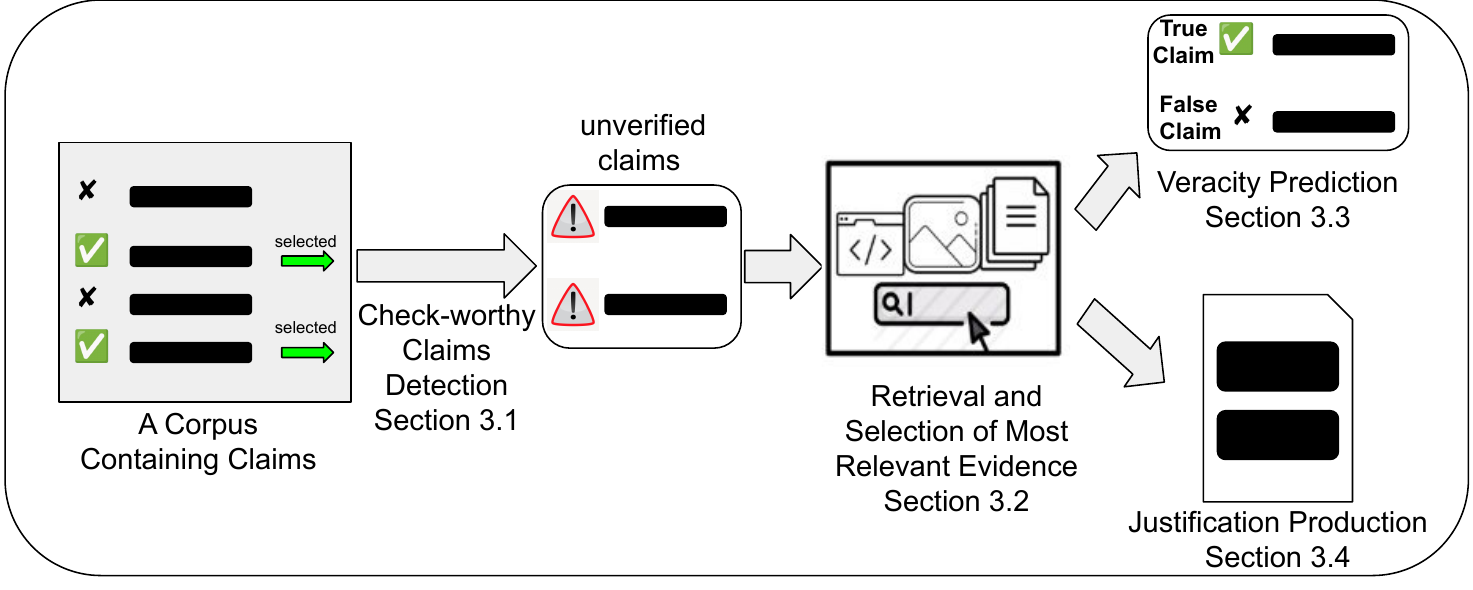}
    \caption{General AFC Pipeline; courtesy of  \cite{guo-etal-2022-survey}.}
    \label{fig:img0}
\end{figure*}

AFC consists of multiple stages forming a pipeline, as shown in Figure \ref{fig:img0}. One of these stages is justification production. In the upcoming subsections, a brief overview of the general stages in the AFC pipeline is provided, with a specific focus on the justification production stage.

\subsection{Check-worthy Claims Detection Stage}
This initial stage classifies the claims as check-worthy or not. If they are check-worthy, then they are selected from the corpus containing them. Deeming if a claim is check-worthy or not is based on the importance of the topic of the claim, if it is verifiable, and if the claim poses potential harm in case it is misleading \cite{guo-etal-2022-survey}.
\subsection{Retrieval and Selection of Most Relevant Evidence}
This stage retrieves data related to the claim from trustworthy sources and selects the most relevant information to make a decision, which is termed the `\textbf{evidence}.' In the subsequent stage, this evidence is used to predict the veracity of the claim. \textbf{The determination of veracity depends on the degree of alignment between the claim and the evidence.} For example, the veracity of the claim `The director of the film `Legend' is English' could depend on the following 
evidence snippets gathered from multiple trustworthy websites: `Brian Helgeland is the director of the film `Legend'', and
`Brian Helgeland only holds a U.S. citizenship'
\subsection{Veracity Prediction of the Claim}
\label{Sec:VP}
This stage classifies claims according to a binary scheme, true or false, or through fine-grained multi-class classification including also other verdicts such as ``partially correct'', or ``correct but misleading without extra context''. Following the example in the previous section, the claim `The director of the film `Legend' is English' should be determined as `False' as it is not aligned with the evidence.

\subsection{Justification Production}
\label{Sec:JP}
This stage produces justifications to explain the verdict of an AFC model regarding a claim's veracity. The process is known as \textbf{justification production} \cite{guo-etal-2022-survey}. 

In the context of the previously discussed claim, `The director of the film `Legend' is English,' an example of a justification for the `False' verdict, grounded in the evidence, could be `Brian Helgeland, the director of the film `Legend,' is American and not English.' Hence, the inputs for a justification production component are the claim and the selected evidence. The veracity verdict may also be an input, depending on the \textbf{pipeline architectures of the AFC systems} that are explained in Section \ref{sec-pipelines} and are shown in our proposed classification of pipelines (see Figure \ref{fig:img1}).

 

We propose categorizing the work in justification production not only based on these pipeline architectures but also on additional dimensions (see Figure \ref{fig:img5}). A key dimension is the \textbf{explainability} of the justification production process. The steps of the process leading to the prediction of the claim's veracity and its justification can be \textbf{\textit{self-explainable or not}}.  

In addition, the \textbf{input type} is an important dimension. It can be either \textbf{\textit{multi-modal}} or \textbf{\textit{text-only}}. Another dimension is the \textbf{nature of the justification output}.
It may be \textbf{\textit{natural language text}}, or just \textbf{\textit{highlighted parts of the input}}, like bold/highlighted words in the claim and evidence, or specific factual \textbf{\textit{triples}} in the form \textbf{\textit{Subject, Predicate, Object (SPO)}} (see Figure \ref{fig:img3} for illustrative examples).

We can also differentiate studies 
based on the \textbf{type of main approaches} utilized, which include:
\textbf{\textit{attention based}} where specific segments of the input having the highest attention scores are highlighted based on the relationship 
between the evidence and the claim;
\textbf{\textit{knowledge graph based}} where a graph is used to represent the evidence. The relevant evidence rationals are selected nodes in the graph, and the edges represent the relations between these selected nodes. Symbolic logic is used to determine if the evidence is aligned with the claim; \textbf{\textit{summarization based}} where the relevant evidence rationals are summarized as natural language text with a focus on whether the claim is aligned with the evidence or not; \textbf{\textit{multi-hop based}} where the claim is decomposed into smaller parts related to each other and these parts are sequentially checked if they are aligned with the evidence or not.; 
and \textbf{\textit{LLMs Retrieval Augmented Generation (RAG)  or Fine-tuning based}} approaches where LLMs are used via prompting to verify the alignment between the claim and the evidence rationals producing the veracity verdict and the justification for the verdict. 

Section \ref{taxo} describes the approaches mentioned above in more detail. The quality of the justifications produced by these approaches is evaluated based on the presence of certain desired properties, which will be discussed in Section \ref{std_just}.

\section{Progression towards Justifications Standardization}
\label{std_just}
\label{counterfact}
%
The aim of justification production is to create a justification that aligns with specific, agreed-upon criteria \cite{sokol2019desiderata}, which we refer to as a \textit{standard justification}. Achieving high-quality justifications involves considering certain desired properties known as \textbf{`desiderata'} \cite{kulesza2015principles}. Researchers have collectively agreed upon these desired properties  \cite{sokol2019desiderata}. Producing justifications aligned with these desired properties is crucial for standardizing justifications in explainable AFC.

\citet{graves2018understanding} identifies key desiderata for justifications, \textbf{completeness}, where the justification must be valid in full contextuality; \textbf{coherence}, ensuring the faithfulness/consistency between the veracity prediction and justification; \textbf{interactivity}, which is putting into consideration the users' feedback; \textbf{actionability}, providing the user with the needed suggestions for modifying the claim to change it from non-factual to factual; \textbf{chronology}, giving preference to the timing of the claim; \textbf{novelty}, ensuring the justification offers new information; \textbf{complexity}, adjusting the justification’s language based on the user’s knowledge; \textbf{parsimony}, favouring more short and concise justifications; \textbf{causality}, where a comprehensive causal model is used for deducing causal connections between inputs and the predictions produced. These properties were defined with further details by \citet{kotonya-toni-2020-explainable}, who also added the desideratum of \textbf{unbiased or impartial justifications}. In the context of fact-checking, bias usually manifests as opinions masquerading as evidence.

\citet{kotonya-toni-2020-explainable-automated} started the first attempt to provide a standard justification evaluation process by measuring two different types of \textbf{coherence} in the produced justifications: 
the \textit{\textbf{global coherence}} which assesses the relevance of a justification in relation to both the claim and its label; and the
\textbf{\textit{local coherence}} which
evaluates the cohesion of sentences within a justification. To maintain local coherence, there should be no contradiction between any two sentences in the justification. \citet{atanasova2022diagnostics} started the first attempt to \textit{generate} standard justifications by adding some desired properties (i.e. \textit{faithfulness}/\textit{coherence}, and \textit{data consistency}) as additional learning signals in the loss function of a transformer-based model
\cite{vaswani2017attention}. 
The \textit{data consistency} evaluates the similarity of justifications for similar input instances.

\section{Datasets in AFC}

It's worth noting that this survey focuses on providing a new taxonomy, a comparative analysis of justification production approaches, investigating pipeline architectures, addressing challenges encountered, and proposing future directions in AFC justification production. Comprehensive examinations of datasets were covered thoroughly in previous surveys (mentioned in Section \ref{r_sur}). However, some information about datasets in AFC is also provided in this section.

The dataset might contain the needed content for all the stages of the fact-checking pipeline: claim detection, evidence retrieval and selection, veracity verdict production, and justification production. 
The following paragraphs will discuss the type of content representing each stage with example datasets provided in Table \ref{tab:table2}. 

Textual \textbf{claims} are the most common input for fact-checking because they are often produced after the claim detection stage. These claims are usually sentence-level statements. Many researchers have created datasets by collecting real-world claims from specialized websites like Politifact because they are easily accessible. Some researchers concentrate on obtaining claims from specialized areas like climate change, science,  and public health. Other sentence-level inputs, such as answers to questions in forums, have also been investigated. Some datasets are English, and some are multi-lingual. Metadata, including information such as publication date, sources, and user profiles, is a common type of \textbf{evidence} considered. Although metadata can provide complementary insights when textual sources or structural knowledge are lacking, it does not directly substantiate the claim. 

Text-based sources, such as news articles, academic publications, and Wikipedia entries, are frequently employed as evidence for fact-checking. Multi-modal claims and evidence have recently been researched, and images and videos are considered more credible than text by most audiences.

Not all the datasets have a binary \textbf{veracity verdict} scheme. Fact-checkers often utilize multi-class labels to classify different levels of truthfulness, including categories such as `true,' `mostly true,' and `mixed.'
\begin{table}[]
\resizebox{\columnwidth}{!}{%
\begin{tabular}{|c|c|c|c|}
\hline
\textbf{Dataset Name} & \textbf{\begin{tabular}[c]{@{}c@{}}Claims\\Number \end{tabular}} & \textbf{\begin{tabular}[c]{@{}c@{}}Veracity\\Verdict\\ Justified\end{tabular}} & \textbf{\begin{tabular}[c]{@{}c@{}}Notes and\\ Remarks\end{tabular}} \\ \hline
\begin{tabular}[c]{@{}c@{}}LIAR \\ \cite{wang-2017-liar}\end{tabular} & 12836 & Yes & \begin{tabular}[c]{@{}c@{}} Political dataset\end{tabular} \\ \hline
\begin{tabular}[c]{@{}c@{}}StatsProperties \\ \cite{vlachos-riedel-2015-identification}\end{tabular} & 7092 & No & \begin{tabular}[c]{@{}c@{}}A knowledge \\graph dataset\end{tabular}  \\ \hline
\begin{tabular}[c]{@{}c@{}}SCIFACT\\ \cite{wadden-etal-2020-fact}\end{tabular} & 1409 & Yes & \begin{tabular}[c]{@{}c@{}}Dataset in the \\scientific domain\end{tabular} \\ \hline
\begin{tabular}[c]{@{}c@{}}MultiFC\\ \cite{augenstein-etal-2019-multifc}\end{tabular} & 36534 & No & \begin{tabular}[c]{@{}c@{}}A multi-domain\\dataset that \\also has metadata\end{tabular} \\ \hline
\begin{tabular}[c]{@{}c@{}}PUBHEALTH\\ \cite{kotonya-toni-2020-explainable-automated}\end{tabular} & 11832 & Yes & \begin{tabular}[c]{@{}c@{}}Dataset in the\\health domain\end{tabular} \\ \hline
\begin{tabular}[c]{@{}c@{}}X-Fact\\ \cite{gupta-srikumar-2021-x}\end{tabular} & 31189 & No & Multi-lingual dataset \\ \hline
\begin{tabular}[c]{@{}c@{}}FEVER\\ \cite{thorne-etal-2018-fever}\end{tabular} & 185445 & No & \begin{tabular}[c]{@{}c@{}}Artificially generated\\ dataset\end{tabular} \\ \hline
\begin{tabular}[c]{@{}c@{}}HOVER\\ \cite{jiang-etal-2020-hover}\end{tabular} & 26171 & No & \begin{tabular}[c]{@{}c@{}}Artificially generated\\ dataset \end{tabular} \\ \hline
\begin{tabular}[c]{@{}c@{}}FEVEROUS\\ \cite{aly-etal-2021-fact}\end{tabular} & 87026 & No & \begin{tabular}[c]{@{}c@{}}Artificially generated\\ dataset and its\\evidence contains\\both text and tables\end{tabular} \\ \hline

\begin{tabular}[c]{@{}c@{}}KLinker\\ \cite{ciampaglia2015computational}\end{tabular} & 10000 & No & \begin{tabular}[c]{@{}c@{}}A knowledge \\ graph dataset \end{tabular} \\ \hline

\begin{tabular}[c]{@{}c@{}}WikiFactCheck\\ \cite{sathe2020automated}\end{tabular} & 124821 & No & \begin{tabular}[c]{@{}c@{}}Artificially generated\\ dataset\end{tabular} \\ \hline



\begin{tabular}[c]{@{}c@{}}VitaminC\\ \cite{schuster-etal-2021-get}\end{tabular} & 488904 & No & \begin{tabular}[c]{@{}c@{}}Artificially generated\\ dataset\end{tabular} \\ \hline
\end{tabular}%
}
\caption{Examples of Datasets for AFC}
\label{tab:table2}
\end{table}

From the perspective of justification production, AFC datasets can be classified into two categories: \textbf{those without justifications} and \textbf{those with justifications}. Complementing datasets lacking justifications is important. This was done by \citet{zhu2023explain} on the HOVER dataset. Table \ref{tab:table2} includes datasets from both categories, with some being synthetically generated and others extracted from external sources like Wikipedia.

\section{Justification Production Taxonomy}
\label{taxooo}
Multiple dimensions or criteria for categorizing \textit{\textbf{Explainable}} Automated Fact-Checking systems are outlined in this section.
We propose five dimensions (illustrated by the first five levels/columns of the Taxonomy tree in Figure \ref{fig:img5}).
The \textbf{Justification Process Explainability} category (Section \ref{explanability})
indicates 
whether the process
leading to 
the justification 
production
is explainable or not. Then 
the \textbf{ Type of Justifications } 
criterion (Section \ref{typee})
indicates
whether 
these are a set of \textit{SOP triples}, \textit{highlighted parts} of selected rationals from the evidence input, or \textit{natural language} textual justifications. Other discriminatory dimensions are the \textbf{Pipeline Architecture} of the AFC components (Section \ref{sec-pipelines}),
the \textbf{Input Type} (Section \ref{multiModal}), whether it is text or multi-modal, and the \textbf{Main Approach} (Section \ref{taxo}), which is the categorization of the predominant methods used for justification production.
In the following sections, every dimension in the taxonomy is discussed in more details.
\subsection{Explainability of Justification Process }
\label{explanability}
The degree of clarity of the process through which the claim is processed and aligned with evidence to produce the justification makes the process self-explanatory. For instance, Multi-hop approaches using QA pairs exemplify this clarity, decomposing claims into parts and then checking their alignment with each evidence snippet. Summarization approaches lack such clarity. For example, consider \textbf{the claim}: ``The director of Interstellar was born in 1960.'' and the corresponding \textbf{evidence snippets}: ``Christopher Nolan was born on 30 July 1970'', ``The name of the director of the film Interstellar is Christopher Nolan.'' and ``Interstellar is a 2014 epic science fiction film.'' The \textbf{justification} of the multi-hop approach \textbf{(self-explainable)} is: "Interstellar is a 2014 science fiction film that was directed by Christopher Nolan. Christopher Nolan was born in 1970, not in 1960, so the claim is false." The \textbf{justification} of the summarization approach \textbf{(non-self-explainable)} is: "Christopher Nolan - born in 1970 - is the director of Interstellar."

In the multi-hop approach, the process of justification production consists of decomposing the claim into three parts: 'Interstellar', 'the director of Interstellar', and 'born in 1960' checking them against relevant evidence snippets. This approach offers clarity by revealing how the input claim is processed and aligned with evidence. In the non-self-explainable justifications, this process is not revealed. This is the case with surveyed summarization approaches. Additionally, the multi-hop justification example provided a \textbf{sequential} explanation. It starts with evidence (Interstellar is a 2014 science fiction film) to address the first part of the claim, i.e., 'Interstellar.' It then proceeds to address the next part of the claim, 'the director of Interstellar', by stating "that was directed by Christopher Nolan,". After that, it addresses the last part of the claim, 'born in 1960', stating that "Christopher Nolan was born in 1970, not in 1960." It outlines a sequence of logical steps grounded in the alignment of claim-evidence to support the conclusion that "the claim is false."

\subsection{Type of Justification}
\label{typee}
The type of justification varies depending on the approach used in the justification production process. Figure \ref{fig:img3} shows examples of different types of output justifications. From Figure \ref{fig:img5}, we can observe that \textit{Natural Language} justifications dominate in recent research as they are the most comprehensible for the readers compared to \textit{SOP triples} or \textit{highlighted words} in the evidence.
\subsection{Justification Production Pipelines Architectures}
\label{sec-pipelines}
\begin{figure*}
    \centering
    \includegraphics[width=1\textwidth, height=10cm]{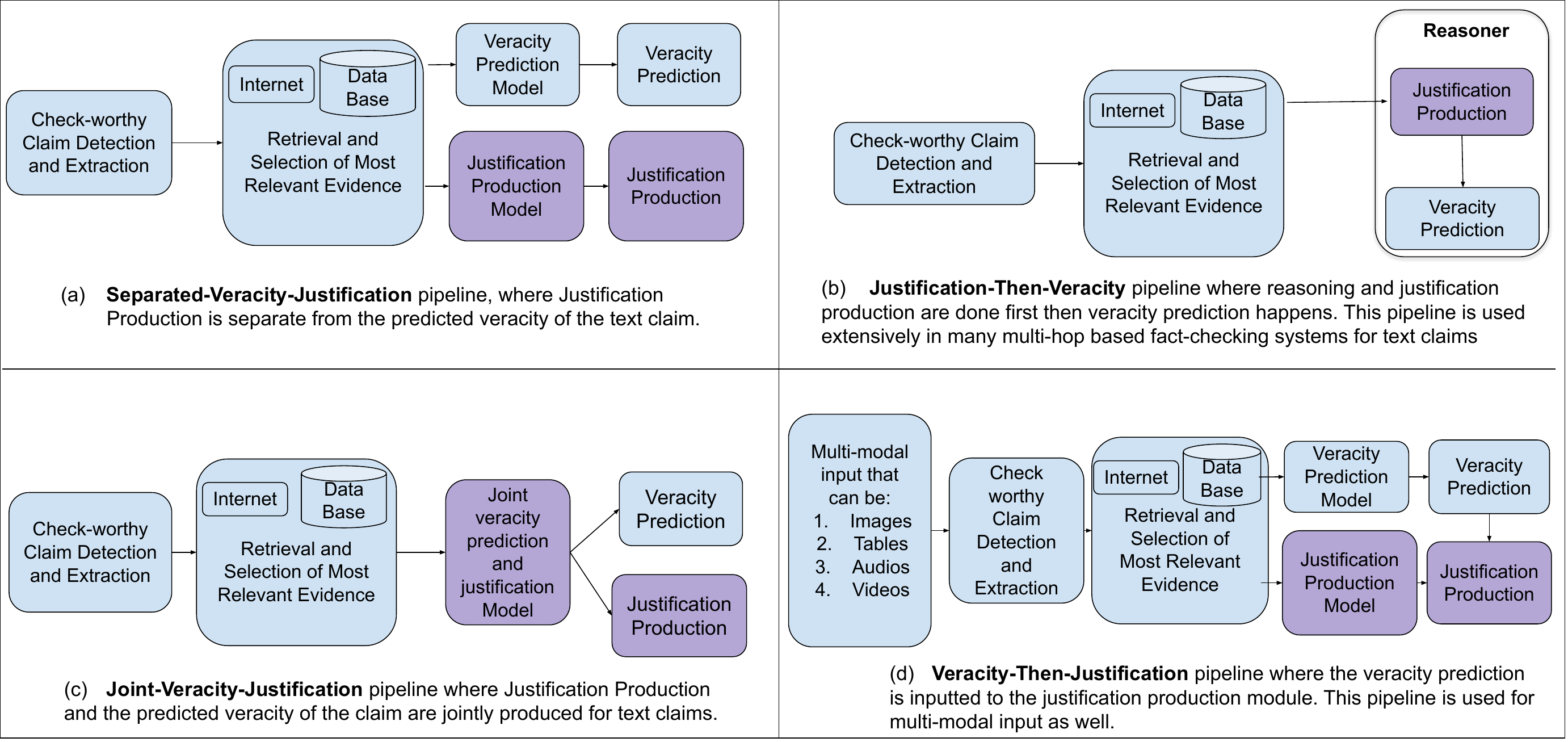}
    \caption{\textbf{The proposed classification of existing pipelines for justification production} based on the type of input (text-only/multi-modal) and the relation between the veracity prediction and the justification production stages.}
    \label{fig:img1}
\end{figure*}

We propose to differentiate various pipelines for Explainable AFC based on the relationship between the justification production stage (Section \ref{Sec:JP}) and the veracity prediction stage (Section \ref{Sec:VP}). These pipelines can be classified into four types, depicted in Figure \ref{fig:img1}. 

In the `\textit{\textbf{Separated-Veracity-Justification}}' pipeline
(Figure \ref{fig:img1}.a), 
the
veracity prediction and justification production are 
independent
processes. This architecture was investigated and used by \citet{atanasova-etal-2020-generating-fact} and \citet{kotonya2020explainable}. 
It is the earliest pipeline offering simpler error tracing capabilities but faces challenges with contradictions between justification and veracity predictions. Research interest in this pipeline is diminishing with the emergence of more robust alternative pipelines like \textit{Justification-Then-Veracity} and \textit{Joint-Veracity-Justification},  as discussed later.

In the `\textit{\textbf{Veracity-Then-Justification}}' pipeline 
(Figure \ref{fig:img1}.d.), 
the veracity 
verdict 
is produced and then inputted
into the justification production module to ensure consistency between the output justification and the verdict in contrast with \textit{Separated-Veracity-Justification}. Moreover, this pipeline allows the usage of different models separately. Each model can handle a different modality; for instance, \citet{yao2023end} trained a sentence-BERT model \cite{reimers2019sentence}  on the textual input while using CLIP \cite{radford2021learning} on the visual input `images.' This pipeline is flexible, allowing a modular design while maintaining consistency between justifications and claim veracity predictions.
It should be noted that this pipeline not only processes multi-modal input but it can also be employed for textual input.

%
%

In the `\textit{\textbf{Joint-Veracity-Justification}}' pipeline (Figure \ref{fig:img1}.c), veracity prediction and justification production are combined tasks carried out by the same model. According to \citet{atanasova-etal-2020-generating-fact}, this pipeline under-performed in summarization compared to the `\textit{Separated-Veracity-Justification}' pipeline. Yet, it excelled in \emph{completeness}, incorporating essential details vital for the fact-checking process. Moreover, it demonstrated superiority in the overall quality of the justifications produced. This pipeline is also used in multi-modal explainable AFC through generating justifications by highlighting the most salient parts of the input having the highest attention scores \cite{kipf2016semi,kou2020exfaux,wu2019mantra, bonettini2021video,purwanto2021fakeclip} as shown in Figure \ref{fig:img5}.

In the `\textit{\textbf{Justification-Then-Veracity}}' pipeline (Figure \ref{fig:img1}.b), a `reasoner' breaks down the claims into smaller segments. It then evaluates each segment of the claim, using available evidence to verify their alignment. Essentially, it employs a logical `AND' operator to determine if all segments of the claim are factual, leading to the final verdict. The verdict is reached after the justification is produced. This pipeline aligns with the most recent research (Figure \ref{fig:img5}). Techniques used in this pipeline include LLMs Chain-of-Thought (CoT) \cite{pan2023fact}, and Multi-hop approaches \cite{wang2023explainable}. \citet{chakraborty-etal-2023-factify3m} uses this pipeline in multi-modal explainable AFC. 
These approaches are further detailed in Section \ref{taxo}.

\subsection{Input Type}
\label{multiModal}

\begin{figure*} 
    \centering
    \includegraphics[width=1\textwidth]{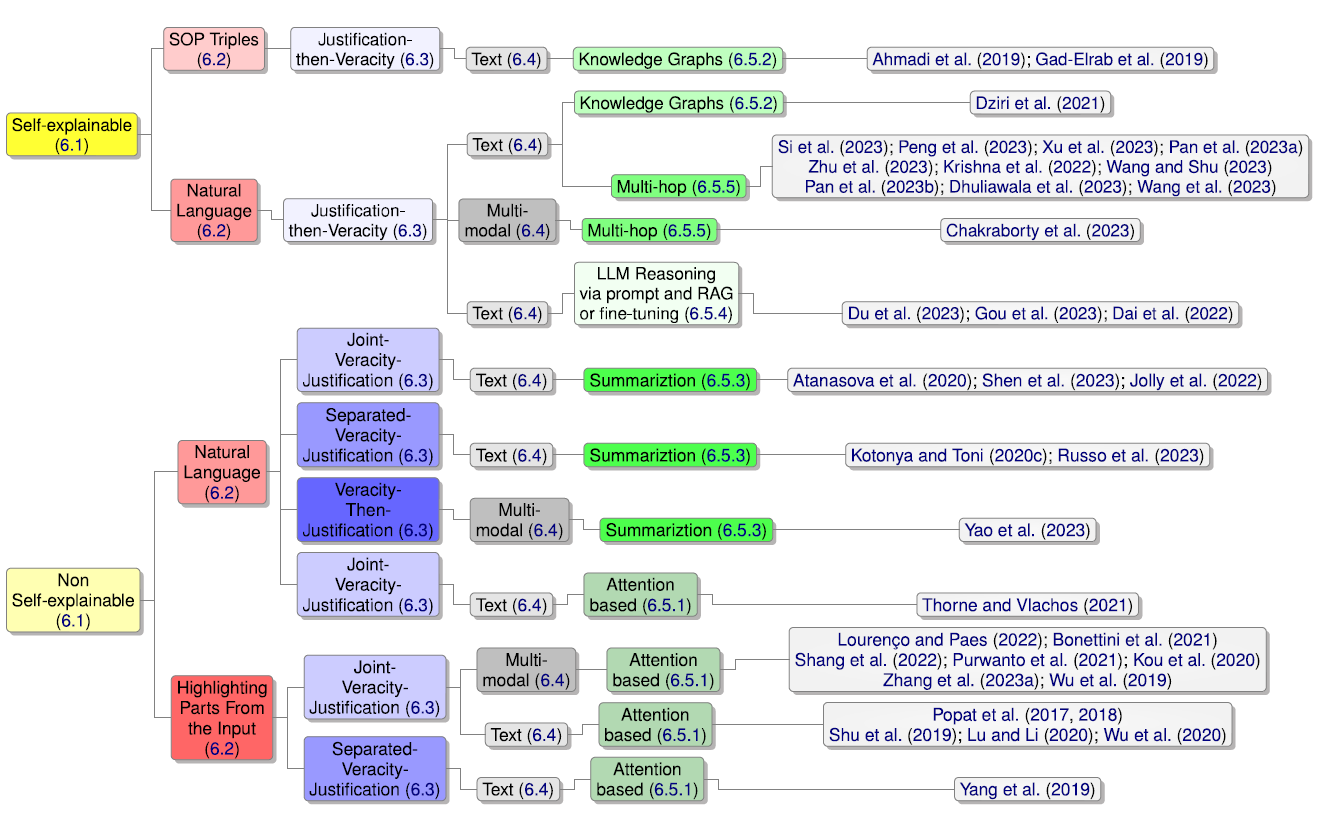}

    \caption{Taxonomy for justification production in AFC according to five dimensions detailed in Section \ref{taxooo}.}
    \label{fig:img5}
\end{figure*}

The input type can be text or multi-modal. The \textbf{text} input is predominant in Explainable AFC. Text-only datasets are more frequent than their multi-modal counterpart (Figure \ref{fig:img5}). 
\textbf{Multi-modal explainable AFC} falls under three categories based on the main approaches dimension: attention based, multi-hop based and summarized natural language text (see Figure \ref{fig:img5}). The attention based approaches like \cite{zhang2023ecenet} have `\textit{Joint-Veracity-Justification}' pipeline architecture, where the input data, like the author of the claim and its timing, are inputted in a fine-tuned transformer based model. Using the attention mechanism, tokens of high attention scores in the evidence and the claim are presented as justification for the veracity verdict. In the new emerging approaches like \cite{yao2023end}, a sentence-BERT is used to process text corpus and a CLIP encoder model is used to present visual features in an image.  All these features are then combined and given to a classifier for verdict prediction and also to BART model \cite{lewis2020bart} for justification production. \citet{chakraborty-etal-2023-factify3m} has used the `\textit{Justification-then-Veracity}' pipeline along with SOTA T5 \cite{raffel2020exploring} for QA pairs generation during claim decomposition and a CNN to analyze visual claims and evidence. Figure \ref{fig:img5} outlines the works
that employ multi-modal input in Explainable AFC, according to the approach involved.



\tikzset{
  my node/.style={
    draw=gray,
    inner color=gray!5,
    outer color=gray!10,
    align=center,
    rounded corners=3,
    text depth=0pt,
    font=\large\sffamily,
    drop shadow,
  }
}

\begin{figure}
\fbox{%
        \begin{minipage}{0.96\columnwidth}
\scriptsize
\textbf{\textcolor{teal}{Claim:}} Earth is Flat.
\\\\
\textbf{\textcolor{teal}{Evidence:}}
    \\ Greeks calculated the radius of the earth thousands of years ago, and it is around 6267 kilometers.
    \\ Nasa images of the earth prove that it is round.
    \\ Boats disappear at large distances, even at sea level and even when we try to use a microscope.
\\\\
\textbf{\textcolor{teal}{Claim's Veracity Verdict:} } False

\begin{center}\textbf{\textcolor{teal}{\fbox{Justifications based on the approaches}}}\end{center}

\textbf{\textcolor{blue}{Attention Based Justification:}}

Greeks calculated the \textit{\textbf{\hl{radius}}} of the earth thousands of years ago and it is around 6267
kilometers.
Nasa images of the earth prove that it is \textbf{\textit{\hl{round}}}.
\\\\
\textbf{\textcolor{blue}{Summarization Based Justification:}}

Nasa images proves that the Earth is  round with a radius of around 6267 km.
\\\\
\textbf{\textcolor{blue}{SOP from a Knowledge Graph Based Approach:}}

LengthOf(Earth radius, 6267 km), Prove(Nasa images,
round earth)
\\\\
\textbf{\textcolor{blue}{Multi-hop Based and Counterfactual Justification:}}

Earth is round as shown by Nasa images. Since it is round, ancient greeks
calculated its radius to be 6267 kms. An interesting fact is that if the Earth was flat, we
would have seen boats even when they are very far using telescopes. However that does not happen due to the curvature of the Earth.
\\\\
\textbf{\textcolor{blue}{LLM Prompting Based Approach using QA pairs:}}

What is Earth? Earth is the planet the we live on.
Is Earth flat? As per Nasa images, No Earth is round.
What is the radius of Earth? As per ancient greeks, it is 6267 km.
What should happen if Earth is flat? Boats would have been seen with strong enough telescopes, disregarding the distances.
 \end{minipage}%
}
    \caption{Illustrative examples of a claim, some evidence snippets that are relevant to the claim, the claim veracity verdict predicted by the model and justifications that could be produced according to different approaches and methodologies.}
    \label{fig:img3}

\end{figure}

\subsection{Main Approaches in Explainable AFC}
\label{taxo}
The following sections detail the main approaches dimension in the taxonomy shown in Figure \ref{fig:img5}. Examples of justifications produced with these approaches are presented in Figure \ref{fig:img3}.
\subsubsection{Attention Based Approaches}
\label{attent}
These approaches mostly use transformer based architectures with attention mechanisms, where justifications are the input segments with the highest attention scores, i.e. justifications are the highest attention score words from the claim and the evidence highlighted in a bold format. As shown in Figure \ref{fig:img5}, they are used with multi-modal input as well as textual input.
The advantage of these approaches is the simplicity of the AFC pipeline compared to the others, as it doesn't have a generator to produce natural language justifications. \citet{thorne-vlachos-2021-evidence}'s work takes a further step by employing a Masked Language Model (MLM) for correcting false claims by replacing their salient false parts with appropriate generated text. However, there are several limitations. \citet{guo-etal-2022-survey} confirm that removing tokens with high attention scores doesn't consistently impact model predictions, questioning the attention mechanisms reliability. Conversely, lower-scoring tokens have been found crucial for accurate predictions. The attention based approaches (Figure \ref{fig:img5}) are categorized into three distinct groups:  (1) \citet{yang2019xfake} employ the `Separated-Veracity-Justification' architecture; (2) \citet{popat2017truth,popat2018declare,shu2019defend,lu-li-2020-gcan,wu-etal-2020-dtca} adopt the `Joint-Veracity-Justification' architecture. Both groups utilize textual input. (3) \citet{lourencco2022modality,bonettini2021video,shang2022duo,purwanto2021fakeclip,kou2020exfaux,wu2019mantra,zhang2023ecenet} address multi-modal input by employing the `Joint-Veracity-Justification' pipeline.

\subsubsection{Knowledge Graph Based Approaches}
\label{KG}
In this approach, justifications are generated based on a graph with all the needed knowledge regarding nodes and relations between these nodes. The computational complexity of a knowledge graph creation from text can limit its scalability. While it provides a structured framework, knowledge graphs may not capture all nuances of natural language. Moreover, they may rely on predefined rules that might not cover all possible scenarios. Additionally, 
the readability of SOP justifications typically produced with this approach might be difficult for non-expert users to comprehend.
Logic rules are needed to search for relevant information in such graphs. For instance, \citet{gad2019exfakt} uses horn rules, which are an implication from an antecedent to a consequent. \citet{ahmadi2019explainable} extended the work by adding probabilistic
answer set programming to the horn rules, while \citet{dziri-etal-2021-neural} used fine-tuned LLMs to traverse the knowledge graphs nodes. 
\subsubsection{Summarization Based Approaches}
\label{summ}
This approach can be extractive, providing short and concise information with less redundancy like in \cite{yao2023end,atanasova-etal-2020-generating-fact,10.1145/3543507.3583375,jolly2022generating} or extractive-abstractive where the extractive summary produced undergoes another process of abstraction. For instance, in the summarization of medical reports, a lot of medical terminology can confuse non-technical audiences, so having a holistic, simpler summary with less technical terminology, such as an `abstractive summary,' is important. This method is implemented in \cite{kotonya2020explainable,russo2023benchmarking}. 
Most of the work on summarization is done using pre-trained models like when \citet{augenstein-etal-2019-multifc} used distilled BERT \cite{sanh2019distilbert}.
\citet{russo2023benchmarking} gave an exhaustive study on enhancing extractive-abstractive summarization, and \citet{jolly2022generating} improved extractive summarization with unsupervised post-editing. The extractive approach lacks the existence of desiderata, while the extractive-abstractive summarization has a higher probability of producing hallucinations than extractive summarization.
\subsubsection{LLMs Reasoning via Prompting and RAG or Fine-tuning}
\label{llms}
LLM prompting, RAG and finetuning are being extensively used as approaches in the domain of explainable AFC. \citet{stammbach2020fever} was among the first researchers to use LLM prompting in explainable AFC. 
Using LLMs generally makes reasoning easier to implement. However, the computational costs are high, and sometimes, LLMs produce hallucinations.
There are many types of hallucinations in LLMs and, in this survey, we focus on fact-conflicting hallucinations.  As per \citet{zhang2023siren}, fact-conflicting hallucinations are produced when LLMs generate information or text that contradicts established world knowledge.

Note that as per \citet{huang2023large}, LLMs can not correct themselves when hallucinating, therefore \citet{gou2023critic} used Chain-of-Thought (CoT) as a possible solution. CoT -introduced by \cite{NEURIPS2022_9d560961}- along with in-context learning and external tools -like search engines-, can greatly reduce hallucinations through reasoning. 

\subsubsection{Multi-hop Approaches}
\label{multi}
Multi-hop approaches are always associated with other methods like graph based methods \cite{xu-etal-2023-counterfactual},
natural logic theorem  \cite{krishna2022proofver}, and QA pair generation along with CoT in LLMs. Multi-hop approaches are being more frequently used in research works like \cite{wang2023explainable,peng2023check,pan2023fact,dhuliawala2023chain,wang2023factcheck,pan2023qacheck}. However, multi-hop fact checking is a complex reasoning task. 
Designing an effective method to generate justifications in the multi-hop setting requires consideration of the logical relationships between the claim and between multiple pieces of evidence. However, the prompts given to LLMs can be enhanced, e.g., using CoT, to exploit more potential of LLMs. 
Generally, Multi-hop LLMs reasoning methods are computationally and financially costly. 

\section{Challenges and Future Directions}
This section outlines the challenges of producing justifications and highlights promising research efforts to address them, suggesting future directions. 

\textbf{Evaluating and Generating Justifications According to Desiderata}
One of the main goals of producing justifications in Explainable AFC is to align them with specific desiderata (Section \ref{counterfact}).
%
Developing quantitative frameworks, or mathematical formulations, is crucial for measuring desiderata in a structured manner. This allows for a systematic comparison of explainable AFC systems based on their incorporation of these desiderata. Furthermore, integrating these measurements into the model training process can significantly enhance justification quality.
%
To date, only \citet{kotonya-toni-2020-explainable-automated} has explored modeling and integrating one particular desired property, coherence/faithfulness, as a learning signal in model training. This area has seen limited further exploration.

Another promising avenue for achieving several desiderata, given the recent proliferation of reasoners with LLMs (Section \ref{llms}) and Multihop (Section \ref{multi}), could be the production of counterfactual justifications, as suggested by \citet{dai2022ask} and illustrated in Figure \ref{fig:img3}. Counterfactual justifications involve imagining scenarios or outcomes that did not actually occur and exploring their consequences. They involve alternate scenarios – for example, `if the Earth were flat, we would be able to see boats even when they are very far away using telescopes' (as shown in Figure \ref{fig:img3}).
When incorporated into the justifications, counterfactual reasoning can reinforce some desired properties such as \textbf{completeness} and \textbf{coherence}. 
Moreover, some desiderata not fully achieved by current work \cite{kotonya-toni-2020-explainable-automated}, like \textbf{actionability}, could also be realized. 
For instance, counterfactual justifications can identify specific elements in a claim that, if altered, could render it factual, thus guiding users toward more accurate statements. 
By offering alternative perspectives on a claim, counterfactual justifications can provide \textbf{novelty} via new information that might not be apparent through traditional justification methods. Counterfactual justifications inherently involve understanding \textbf{causal} relationships, another highly desired property. The ultimate objective remains to incorporate most or all of the desiderata presented in Section \ref{std_just}.

\textbf{Natural Language Justifications in Multi-modal AFC} 
%
The majority of works processing multi-modal input rely on attention-based approaches 
(Section \ref{fig:img5}). Commonly, these works use highlighted input segments as justifications. However, such justifications are less effective in meeting the desired properties compared to valid natural language. Only a few recent studies have incorporated multi-modality in the input while producing natural language justifications: \citet{yao2023end}
and \citet{chakraborty-etal-2023-factify3m}.
\citet{yao2023end} use a `\textit{Veracity-Then-Justification}' process.
However, this method is less intuitive 
compared to the much more popular `\textit{Justification-Then-Veracity}' pipeline. A key feature of this latter architecture is the inclusion of a reasoning component, which attempts to deduce veracity based on justifications grounded in the evidence. It is, however, predominantly used with text-only inputs.  \citet{chakraborty-etal-2023-factify3m} are the only researchers so far to use this architecture with multi-modal input (Section \ref{multiModal}).
There is potential for further research in this area, especially with the advancements in LLMs that can process multi-modal inputs and produce coherent, natural language justifications, similar to the approach used by \citet{lin2024towards} in a different context of generating explanations for identifying harmful content in memes.

\textbf{Non-factual Hallucinations in LLMs in AFC} Nowadays, LLMs are used more frequently in AFC. The challenge is that they themselves can produce hallucinations. There are many types of hallucinations; however, the most related type to the domain of AFC is non-factual hallucinations. 
Aiming to address hallucinations, \citet{du2023improving} introduced the Society of Minds (SOM) to improve the factuality and accuracy of the LLMs output. SOM is a method where multiple instances of the same language model produce results for the same query, and then they debate to unify and improve their answers, correcting hallucinations in multiple rounds. CoT is also used during these rounds. This method is based on the hypothesis that hallucinations are not produced consistently by LLMs. The debate rounds can also happen between different models like chatGPT versus BARD \cite{ahmed2023chatgpt}.

\textbf{Complexity of Justification Production} Generally, justification production via Multi-hop or LLMs reasoning methods are computationally and financially costly. For instance, employing FOLK \cite{pan2023fact} led to an expense of 20 USD for every 100 examples when using the OpenAI API, or required 7.5 hours of processing time on locally deployed llama-30B models with an 8x A5000 cluster. Addressing the computational cost associated with reasoning methods justification production is essential, warranting exploration of techniques such as knowledge distillation \cite{hinton2015distilling} and quantization \cite{choukroun2019low}.

\section{Conclusion}

In summary, this survey contributes a novel multi-dimensional taxonomy, comprehensively presents the architectures employed in justification production, explores emergent methodologies, conducts a comparative analysis of these methodologies, and proposes prospective avenues for further research. 






\section*{Limitations}
The limitations in this survey can be summarized in the following points:
\begin{enumerate}
    \item 
    We have not included work on AFC that focuses solely on claim verification based on the language and lexicons used in the claims.
    \item The few related papers that were published before 2015 were not included in the taxonomy.
    \item This survey focused only on the work on English justification production. Multi-lingual justification production should also be explored. 
    
\end{enumerate}




\bibliography{anthology,custom}

\begin{thebibliography}{76}
\expandafter\ifx\csname natexlab\endcsname\relax\def\natexlab#1{#1}\fi

\bibitem[{Ahmadi et~al.(2019)Ahmadi, Lee, Papotti, and Saeed}]{ahmadi2019explainable}
Naser Ahmadi, Joohyung Lee, Paolo Papotti, and Mohammed Saeed. 2019.
\newblock Explainable fact checking with probabilistic answer set programming.
\newblock \emph{arXiv preprint arXiv:1906.09198}.

\bibitem[{Ahmed et~al.(2023)Ahmed, Roy, Kajol, Hasan, Datta, and Reza}]{ahmed2023chatgpt}
Imtiaz Ahmed, Ayon Roy, Mashrafi Kajol, Uzma Hasan, Partha~Protim Datta, and Md~Rokonuzzaman Reza. 2023.
\newblock Chatgpt vs. bard: a comparative study.
\newblock \emph{Authorea Preprints}.

\bibitem[{Alam et~al.(2022)Alam, Cresci, Chakraborty, Silvestri, Dimitrov, Martino, Shaar, Firooz, and Nakov}]{alam-etal-2022-survey}
Firoj Alam, Stefano Cresci, Tanmoy Chakraborty, Fabrizio Silvestri, Dimiter Dimitrov, Giovanni Da~San Martino, Shaden Shaar, Hamed Firooz, and Preslav Nakov. 2022.
\newblock \href {https://aclanthology.org/2022.coling-1.576} {A survey on multimodal disinformation detection}.
\newblock In \emph{Proceedings of the 29th International Conference on Computational Linguistics}, pages 6625--6643, Gyeongju, Republic of Korea. International Committee on Computational Linguistics.

\bibitem[{Aly et~al.(2021)Aly, Guo, Schlichtkrull, Thorne, Vlachos, Christodoulopoulos, Cocarascu, and Mittal}]{aly-etal-2021-fact}
Rami Aly, Zhijiang Guo, Michael~Sejr Schlichtkrull, James Thorne, Andreas Vlachos, Christos Christodoulopoulos, Oana Cocarascu, and Arpit Mittal. 2021.
\newblock \href {https://doi.org/10.18653/v1/2021.fever-1.1} {The fact extraction and {VER}ification over unstructured and structured information ({FEVEROUS}) shared task}.
\newblock In \emph{Proceedings of the Fourth Workshop on Fact Extraction and VERification (FEVER)}, pages 1--13, Dominican Republic. Association for Computational Linguistics.

\bibitem[{Atanasova et~al.(2020)Atanasova, Simonsen, Lioma, and Augenstein}]{atanasova-etal-2020-generating-fact}
Pepa Atanasova, Jakob~Grue Simonsen, Christina Lioma, and Isabelle Augenstein. 2020.
\newblock \href {https://doi.org/10.18653/v1/2020.acl-main.656} {Generating fact checking explanations}.
\newblock In \emph{Proceedings of the 58th Annual Meeting of the Association for Computational Linguistics}, pages 7352--7364, Online. Association for Computational Linguistics.

\bibitem[{Atanasova et~al.(2022)Atanasova, Simonsen, Lioma, and Augenstein}]{atanasova2022diagnostics}
Pepa Atanasova, Jakob~Grue Simonsen, Christina Lioma, and Isabelle Augenstein. 2022.
\newblock Diagnostics-guided explanation generation.
\newblock In \emph{Proceedings of the AAAI Conference on Artificial Intelligence}, volume~36, pages 10445--10453.

\bibitem[{Augenstein et~al.(2019)Augenstein, Lioma, Wang, Chaves~Lima, Hansen, Hansen, and Simonsen}]{augenstein-etal-2019-multifc}
Isabelle Augenstein, Christina Lioma, Dongsheng Wang, Lucas Chaves~Lima, Casper Hansen, Christian Hansen, and Jakob~Grue Simonsen. 2019.
\newblock \href {https://doi.org/10.18653/v1/D19-1475} {{M}ulti{FC}: A real-world multi-domain dataset for evidence-based fact checking of claims}.
\newblock In \emph{Proceedings of the 2019 Conference on Empirical Methods in Natural Language Processing and the 9th International Joint Conference on Natural Language Processing (EMNLP-IJCNLP)}, pages 4685--4697, Hong Kong, China. Association for Computational Linguistics.

\bibitem[{Bonettini et~al.(2021)Bonettini, Cannas, Mandelli, Bondi, Bestagini, and Tubaro}]{bonettini2021video}
Nicolo Bonettini, Edoardo~Daniele Cannas, Sara Mandelli, Luca Bondi, Paolo Bestagini, and Stefano Tubaro. 2021.
\newblock Video face manipulation detection through ensemble of cnns.
\newblock In \emph{2020 25th international conference on pattern recognition (ICPR)}, pages 5012--5019. IEEE.

\bibitem[{Chakraborty et~al.(2023)Chakraborty, Pahwa, Rani, Chatterjee, Dalal, Dave, G, Gurumurthy, Mahor, Mukherjee, Pakala, Paul, Reddy, Sarkar, Sensharma, Chadha, Sheth, and Das}]{chakraborty-etal-2023-factify3m}
Megha Chakraborty, Khushbu Pahwa, Anku Rani, Shreyas Chatterjee, Dwip Dalal, Harshit Dave, Ritvik G, Preethi Gurumurthy, Adarsh Mahor, Samahriti Mukherjee, Aditya Pakala, Ishan Paul, Janvita Reddy, Arghya Sarkar, Kinjal Sensharma, Aman Chadha, Amit Sheth, and Amitava Das. 2023.
\newblock \href {https://doi.org/10.18653/v1/2023.emnlp-main.945} {{FACTIFY}3{M}: A benchmark for multimodal fact verification with explainability through 5{W} question-answering}.
\newblock In \emph{Proceedings of the 2023 Conference on Empirical Methods in Natural Language Processing}, pages 15282--15322, Singapore. Association for Computational Linguistics.

\bibitem[{Choukroun et~al.(2019)Choukroun, Kravchik, Yang, and Kisilev}]{choukroun2019low}
Yoni Choukroun, Eli Kravchik, Fan Yang, and Pavel Kisilev. 2019.
\newblock Low-bit quantization of neural networks for efficient inference.
\newblock In \emph{2019 IEEE/CVF International Conference on Computer Vision Workshop (ICCVW)}, pages 3009--3018. IEEE.

\bibitem[{Ciampaglia et~al.(2015)Ciampaglia, Shiralkar, Rocha, Bollen, Menczer, and Flammini}]{ciampaglia2015computational}
Giovanni~Luca Ciampaglia, Prashant Shiralkar, Luis~M Rocha, Johan Bollen, Filippo Menczer, and Alessandro Flammini. 2015.
\newblock Computational fact checking from knowledge networks.
\newblock \emph{PloS one}, 10(6):e0128193.

\bibitem[{Dai et~al.(2022)Dai, Hsu, Xiong, and Ku}]{dai2022ask}
Shih-Chieh Dai, Yi-Li Hsu, Aiping Xiong, and Lun-Wei Ku. 2022.
\newblock Ask to know more: Generating counterfactual explanations for fake claims.
\newblock In \emph{Proceedings of the 28th ACM SIGKDD Conference on Knowledge Discovery and Data Mining}, pages 2800--2810.

\bibitem[{Dhuliawala et~al.(2023)Dhuliawala, Komeili, Xu, Raileanu, Li, Celikyilmaz, and Weston}]{dhuliawala2023chain}
Shehzaad Dhuliawala, Mojtaba Komeili, Jing Xu, Roberta Raileanu, Xian Li, Asli Celikyilmaz, and Jason Weston. 2023.
\newblock Chain-of-verification reduces hallucination in large language models.
\newblock \emph{arXiv preprint arXiv:2309.11495}.

\bibitem[{Du et~al.(2023)Du, Li, Torralba, Tenenbaum, and Mordatch}]{du2023improving}
Yilun Du, Shuang Li, Antonio Torralba, Joshua~B Tenenbaum, and Igor Mordatch. 2023.
\newblock Improving factuality and reasoning in language models through multiagent debate.
\newblock \emph{arXiv preprint arXiv:2305.14325}.

\bibitem[{Dziri et~al.(2021)Dziri, Madotto, Za{\"\i}ane, and Bose}]{dziri-etal-2021-neural}
Nouha Dziri, Andrea Madotto, Osmar Za{\"\i}ane, and Avishek~Joey Bose. 2021.
\newblock \href {https://doi.org/10.18653/v1/2021.emnlp-main.168} {Neural path hunter: Reducing hallucination in dialogue systems via path grounding}.
\newblock In \emph{Proceedings of the 2021 Conference on Empirical Methods in Natural Language Processing}, pages 2197--2214, Online and Punta Cana, Dominican Republic. Association for Computational Linguistics.

\bibitem[{Gad-Elrab et~al.(2019)Gad-Elrab, Stepanova, Urbani, and Weikum}]{gad2019exfakt}
Mohamed~H Gad-Elrab, Daria Stepanova, Jacopo Urbani, and Gerhard Weikum. 2019.
\newblock Exfakt: A framework for explaining facts over knowledge graphs and text.
\newblock In \emph{Proceedings of the twelfth ACM international conference on web search and data mining}, pages 87--95.

\bibitem[{Gou et~al.(2023)Gou, Shao, Gong, Shen, Yang, Duan, and Chen}]{gou2023critic}
Zhibin Gou, Zhihong Shao, Yeyun Gong, Yelong Shen, Yujiu Yang, Nan Duan, and Weizhu Chen. 2023.
\newblock Critic: Large language models can self-correct with tool-interactive critiquing.
\newblock \emph{arXiv preprint arXiv:2305.11738}.

\bibitem[{Graves(2018)}]{graves2018understanding}
D~Graves. 2018.
\newblock Understanding the promise and limits of automated fact-checking.
\newblock \emph{Reuters Institute for the Study of Journalism}.

\bibitem[{Guo et~al.(2022)Guo, Schlichtkrull, and Vlachos}]{guo-etal-2022-survey}
Zhijiang Guo, Michael Schlichtkrull, and Andreas Vlachos. 2022.
\newblock \href {https://doi.org/10.1162/tacl_a_00454} {A survey on automated fact-checking}.
\newblock \emph{Transactions of the Association for Computational Linguistics}, 10:178--206.

\bibitem[{Gupta and Srikumar(2021)}]{gupta-srikumar-2021-x}
Ashim Gupta and Vivek Srikumar. 2021.
\newblock \href {https://doi.org/10.18653/v1/2021.acl-short.86} {{X}-fact: A new benchmark dataset for multilingual fact checking}.
\newblock In \emph{Proceedings of the 59th Annual Meeting of the Association for Computational Linguistics and the 11th International Joint Conference on Natural Language Processing (Volume 2: Short Papers)}, pages 675--682, Online. Association for Computational Linguistics.

\bibitem[{Hinton et~al.(2015)Hinton, Vinyals, and Dean}]{hinton2015distilling}
Geoffrey Hinton, Oriol Vinyals, and Jeff Dean. 2015.
\newblock Distilling the knowledge in a neural network.
\newblock \emph{stat}, 1050:9.

\bibitem[{Huang et~al.(2023)Huang, Chen, Mishra, Zheng, Yu, Song, and Zhou}]{huang2023large}
Jie Huang, Xinyun Chen, Swaroop Mishra, Huaixiu~Steven Zheng, Adams~Wei Yu, Xinying Song, and Denny Zhou. 2023.
\newblock Large language models cannot self-correct reasoning yet.
\newblock \emph{arXiv preprint arXiv:2310.01798}.

\bibitem[{Jiang et~al.(2020)Jiang, Bordia, Zhong, Dognin, Singh, and Bansal}]{jiang-etal-2020-hover}
Yichen Jiang, Shikha Bordia, Zheng Zhong, Charles Dognin, Maneesh Singh, and Mohit Bansal. 2020.
\newblock \href {https://doi.org/10.18653/v1/2020.findings-emnlp.309} {{H}o{V}er: A dataset for many-hop fact extraction and claim verification}.
\newblock In \emph{Findings of the Association for Computational Linguistics: EMNLP 2020}, pages 3441--3460, Online. Association for Computational Linguistics.

\bibitem[{Jolly et~al.(2022)Jolly, Atanasova, and Augenstein}]{jolly2022generating}
Shailza Jolly, Pepa Atanasova, and Isabelle Augenstein. 2022.
\newblock Generating fluent fact checking explanations with unsupervised post-editing.
\newblock \emph{Information}, 13(10):500.

\bibitem[{Kipf and Welling(2016)}]{kipf2016semi}
Thomas~N Kipf and Max Welling. 2016.
\newblock Semi-supervised classification with graph convolutional networks.
\newblock \emph{arXiv preprint arXiv:1609.02907}.

\bibitem[{Kotonya and Toni(2020{\natexlab{a}})}]{kotonya-toni-2020-explainable}
Neema Kotonya and Francesca Toni. 2020{\natexlab{a}}.
\newblock \href {https://doi.org/10.18653/v1/2020.coling-main.474} {Explainable automated fact-checking: A survey}.
\newblock In \emph{Proceedings of the 28th International Conference on Computational Linguistics}, pages 5430--5443, Barcelona, Spain (Online). International Committee on Computational Linguistics.

\bibitem[{Kotonya and Toni(2020{\natexlab{b}})}]{kotonya-toni-2020-explainable-automated}
Neema Kotonya and Francesca Toni. 2020{\natexlab{b}}.
\newblock \href {https://doi.org/10.18653/v1/2020.emnlp-main.623} {Explainable automated fact-checking for public health claims}.
\newblock In \emph{Proceedings of the 2020 Conference on Empirical Methods in Natural Language Processing (EMNLP)}, pages 7740--7754, Online. Association for Computational Linguistics.

\bibitem[{Kotonya and Toni(2020{\natexlab{c}})}]{kotonya2020explainable}
Neema Kotonya and Francesca Toni. 2020{\natexlab{c}}.
\newblock Explainable automated fact-checking for public health claims.
\newblock \emph{arXiv preprint arXiv:2010.09926}.

\bibitem[{Kou et~al.(2020)Kou, Zhang, Shang, and Wang}]{kou2020exfaux}
Ziyi Kou, Daniel~Yue Zhang, Lanyu Shang, and Dong Wang. 2020.
\newblock Exfaux: A weakly supervised approach to explainable fauxtography detection.
\newblock In \emph{2020 IEEE International Conference on Big Data (Big Data)}, pages 631--636. IEEE.

\bibitem[{Krishna et~al.(2022)Krishna, Riedel, and Vlachos}]{krishna2022proofver}
Amrith Krishna, Sebastian Riedel, and Andreas Vlachos. 2022.
\newblock Proofver: Natural logic theorem proving for fact verification.
\newblock \emph{Transactions of the Association for Computational Linguistics}, 10:1013--1030.

\bibitem[{Kulesza et~al.(2015)Kulesza, Burnett, Wong, and Stumpf}]{kulesza2015principles}
Todd Kulesza, Margaret Burnett, Weng-Keen Wong, and Simone Stumpf. 2015.
\newblock Principles of explanatory debugging to personalize interactive machine learning.
\newblock In \emph{Proceedings of the 20th international conference on intelligent user interfaces}, pages 126--137.

\bibitem[{Lewandowsky et~al.(2012)Lewandowsky, Ecker, Seifert, Schwarz, and Cook}]{lewandowsky2012misinformation}
Stephan Lewandowsky, Ullrich~KH Ecker, Colleen~M Seifert, Norbert Schwarz, and John Cook. 2012.
\newblock Misinformation and its correction: Continued influence and successful debiasing.
\newblock \emph{Psychological science in the public interest}, 13(3):106--131.

\bibitem[{Lewis et~al.(2020)Lewis, Liu, Goyal, Ghazvininejad, Mohamed, Levy, Stoyanov, and Zettlemoyer}]{lewis2020bart}
Mike Lewis, Yinhan Liu, Naman Goyal, Marjan Ghazvininejad, Abdelrahman Mohamed, Omer Levy, Veselin Stoyanov, and Luke Zettlemoyer. 2020.
\newblock Bart: Denoising sequence-to-sequence pre-training for natural language generation, translation, and comprehension.
\newblock In \emph{Proceedings of the 58th Annual Meeting of the Association for Computational Linguistics}, pages 7871--7880.

\bibitem[{Lin et~al.(2024)Lin, Luo, Gao, Ma, Wang, and Yang}]{lin2024towards}
Hongzhan Lin, Ziyang Luo, Wei Gao, Jing Ma, Bo~Wang, and Ruichao Yang. 2024.
\newblock Towards explainable harmful meme detection through multimodal debate between large language models.
\newblock \emph{arXiv preprint arXiv:2401.13298}.

\bibitem[{Louren{\c{c}}o and Paes(2022)}]{lourencco2022modality}
V{\'\i}tor Louren{\c{c}}o and Aline Paes. 2022.
\newblock A modality-level explainable framework for misinformation checking in social networks.
\newblock \emph{arXiv preprint arXiv:2212.04272}.

\bibitem[{Lu and Li(2020)}]{lu-li-2020-gcan}
Yi-Ju Lu and Cheng-Te Li. 2020.
\newblock \href {https://doi.org/10.18653/v1/2020.acl-main.48} {{GCAN}: Graph-aware co-attention networks for explainable fake news detection on social media}.
\newblock In \emph{Proceedings of the 58th Annual Meeting of the Association for Computational Linguistics}, pages 505--514, Online. Association for Computational Linguistics.

\bibitem[{Nakov et~al.(2021{\natexlab{a}})Nakov, Corney, Hasanain, Alam, Elsayed, Barron-Cedeno, Papotti, Shaar, Da~San~Martino et~al.}]{nakov2021automated}
P~Nakov, D~Corney, M~Hasanain, F~Alam, T~Elsayed, A~Barron-Cedeno, P~Papotti, S~Shaar, G~Da~San~Martino, et~al. 2021{\natexlab{a}}.
\newblock Automated fact-checking for assisting human fact-checkers.
\newblock In \emph{IJCAI}, pages 4551--4558. International Joint Conferences on Artificial Intelligence.

\bibitem[{Nakov et~al.(2021{\natexlab{b}})Nakov, Corney, Hasanain, Alam, Elsayed, Barrón-Cedeño, Papotti, Shaar, and Da~San~Martino}]{ijcai2021p619}
Preslav Nakov, David Corney, Maram Hasanain, Firoj Alam, Tamer Elsayed, Alberto Barrón-Cedeño, Paolo Papotti, Shaden Shaar, and Giovanni Da~San~Martino. 2021{\natexlab{b}}.
\newblock \href {https://doi.org/10.24963/ijcai.2021/619} {Automated fact-checking for assisting human fact-checkers}.
\newblock In \emph{Proceedings of the Thirtieth International Joint Conference on Artificial Intelligence, {IJCAI-21}}, pages 4551--4558. International Joint Conferences on Artificial Intelligence Organization.
\newblock Survey Track.

\bibitem[{Pan et~al.(2023{\natexlab{a}})Pan, Lu, Kan, and Nakov}]{pan2023qacheck}
Liangming Pan, Xinyuan Lu, Min-Yen Kan, and Preslav Nakov. 2023{\natexlab{a}}.
\newblock Qacheck: A demonstration system for question-guided multi-hop fact-checking.
\newblock \emph{arXiv preprint arXiv:2310.07609}.

\bibitem[{Pan et~al.(2023{\natexlab{b}})Pan, Wu, Lu, Luu, Wang, Kan, and Nakov}]{pan2023fact}
Liangming Pan, Xiaobao Wu, Xinyuan Lu, Anh~Tuan Luu, William~Yang Wang, Min-Yen Kan, and Preslav Nakov. 2023{\natexlab{b}}.
\newblock Fact-checking complex claims with program-guided reasoning.
\newblock \emph{arXiv preprint arXiv:2305.12744}.

\bibitem[{Peng et~al.(2023)Peng, Galley, He, Cheng, Xie, Hu, Huang, Liden, Yu, Chen et~al.}]{peng2023check}
Baolin Peng, Michel Galley, Pengcheng He, Hao Cheng, Yujia Xie, Yu~Hu, Qiuyuan Huang, Lars Liden, Zhou Yu, Weizhu Chen, et~al. 2023.
\newblock Check your facts and try again: Improving large language models with external knowledge and automated feedback.
\newblock \emph{arXiv preprint arXiv:2302.12813}.

\bibitem[{Popat et~al.(2017)Popat, Mukherjee, Str{\"o}tgen, and Weikum}]{popat2017truth}
Kashyap Popat, Subhabrata Mukherjee, Jannik Str{\"o}tgen, and Gerhard Weikum. 2017.
\newblock Where the truth lies: Explaining the credibility of emerging claims on the web and social media.
\newblock In \emph{Proceedings of the 26th International Conference on World Wide Web Companion}, pages 1003--1012.

\bibitem[{Popat et~al.(2018)Popat, Mukherjee, Yates, and Weikum}]{popat2018declare}
Kashyap Popat, Subhabrata Mukherjee, Andrew Yates, and Gerhard Weikum. 2018.
\newblock Declare: Debunking fake news and false claims using evidence-aware deep learning.
\newblock \emph{arXiv preprint arXiv:1809.06416}.

\bibitem[{Purwanto et~al.(2021)Purwanto, Santoso, Lei, Yang, and Peng}]{purwanto2021fakeclip}
Christian~Nathaniel Purwanto, Joan Santoso, Po-Ruey Lei, Hui-Kuo Yang, and Wen-Chih Peng. 2021.
\newblock Fakeclip: Multimodal fake caption detection with mixed languages for explainable visualization.
\newblock In \emph{2021 International Conference on Technologies and Applications of Artificial Intelligence (TAAI)}, pages 1--6. IEEE.

\bibitem[{Radford et~al.(2021)Radford, Kim, Hallacy, Ramesh, Goh, Agarwal, Sastry, Askell, Mishkin, Clark et~al.}]{radford2021learning}
Alec Radford, Jong~Wook Kim, Chris Hallacy, Aditya Ramesh, Gabriel Goh, Sandhini Agarwal, Girish Sastry, Amanda Askell, Pamela Mishkin, Jack Clark, et~al. 2021.
\newblock Learning transferable visual models from natural language supervision.
\newblock In \emph{International conference on machine learning}, pages 8748--8763. PMLR.

\bibitem[{Raffel et~al.(2020)Raffel, Shazeer, Roberts, Lee, Narang, Matena, Zhou, Li, and Liu}]{raffel2020exploring}
Colin Raffel, Noam Shazeer, Adam Roberts, Katherine Lee, Sharan Narang, Michael Matena, Yanqi Zhou, Wei Li, and Peter~J Liu. 2020.
\newblock Exploring the limits of transfer learning with a unified text-to-text transformer.
\newblock \emph{The Journal of Machine Learning Research}, 21(1):5485--5551.

\bibitem[{Reimers and Gurevych(2019)}]{reimers2019sentence}
Nils Reimers and Iryna Gurevych. 2019.
\newblock Sentence-bert: Sentence embeddings using siamese bert-networks.
\newblock In \emph{Proceedings of the 2019 Conference on Empirical Methods in Natural Language Processing and the 9th International Joint Conference on Natural Language Processing (EMNLP-IJCNLP)}. Association for Computational Linguistics.

\bibitem[{Russo et~al.(2023)Russo, Tekiro{\u{g}}lu, and Guerini}]{russo2023benchmarking}
Daniel Russo, Serra~Sinem Tekiro{\u{g}}lu, and Marco Guerini. 2023.
\newblock Benchmarking the generation of fact checking explanations.
\newblock \emph{Transactions of the Association for Computational Linguistics}, 11:1250--1264.

\bibitem[{Sanh et~al.(2019)Sanh, Debut, Chaumond, and Wolf}]{sanh2019distilbert}
Victor Sanh, Lysandre Debut, Julien Chaumond, and Thomas Wolf. 2019.
\newblock Distilbert, a distilled version of bert: smaller, faster, cheaper and lighter.
\newblock \emph{arXiv preprint arXiv:1910.01108}.

\bibitem[{Sathe et~al.(2020)Sathe, Ather, Le, Perry, and Park}]{sathe2020automated}
Aalok Sathe, Salar Ather, Tuan~Manh Le, Nathan Perry, and Joonsuk Park. 2020.
\newblock Automated fact-checking of claims from wikipedia.
\newblock In \emph{Proceedings of the Twelfth Language Resources and Evaluation Conference}, pages 6874--6882.

\bibitem[{Schuster et~al.(2021)Schuster, Fisch, and Barzilay}]{schuster-etal-2021-get}
Tal Schuster, Adam Fisch, and Regina Barzilay. 2021.
\newblock \href {https://doi.org/10.18653/v1/2021.naacl-main.52} {Get your vitamin {C}! robust fact verification with contrastive evidence}.
\newblock In \emph{Proceedings of the 2021 Conference of the North American Chapter of the Association for Computational Linguistics: Human Language Technologies}, pages 624--643, Online. Association for Computational Linguistics.

\bibitem[{Shang et~al.(2022)Shang, Kou, Zhang, and Wang}]{shang2022duo}
Lanyu Shang, Ziyi Kou, Yang Zhang, and Dong Wang. 2022.
\newblock A duo-generative approach to explainable multimodal covid-19 misinformation detection.
\newblock In \emph{Proceedings of the ACM Web Conference 2022}, pages 3623--3631.

\bibitem[{Shen et~al.(2023)Shen, Liu, Finnie, Rahmati, Bendersky, and Najork}]{10.1145/3543507.3583375}
Jiaming Shen, Jialu Liu, Dan Finnie, Negar Rahmati, Mike Bendersky, and Marc Najork. 2023.
\newblock \href {https://doi.org/10.1145/3543507.3583375} {“why is this misleading?”: Detecting news headline hallucinations with explanations}.
\newblock In \emph{Proceedings of the ACM Web Conference 2023}, WWW '23, page 1662–1672, New York, NY, USA. Association for Computing Machinery.

\bibitem[{Shu et~al.(2019)Shu, Cui, Wang, Lee, and Liu}]{shu2019defend}
Kai Shu, Limeng Cui, Suhang Wang, Dongwon Lee, and Huan Liu. 2019.
\newblock defend: Explainable fake news detection.
\newblock In \emph{Proceedings of the 25th ACM SIGKDD international conference on knowledge discovery \& data mining}, pages 395--405.

\bibitem[{Sokol and Flach(2019)}]{sokol2019desiderata}
Kacper Sokol and Peter Flach. 2019.
\newblock Desiderata for interpretability: explaining decision tree predictions with counterfactuals.
\newblock In \emph{Proceedings of the AAAI conference on artificial intelligence}, volume~33, pages 10035--10036.

\bibitem[{Stammbach and Ash(2020)}]{stammbach2020fever}
Dominik Stammbach and Elliott Ash. 2020.
\newblock e-fever: Explanations and summaries for automated fact checking.
\newblock \emph{Proceedings of the 2020 Truth and Trust Online (TTO 2020)}, pages 32--43.

\bibitem[{Thorne and Vlachos(2018)}]{thorne-vlachos-2018-automated}
James Thorne and Andreas Vlachos. 2018.
\newblock \href {https://aclanthology.org/C18-1283} {Automated fact checking: Task formulations, methods and future directions}.
\newblock In \emph{Proceedings of the 27th International Conference on Computational Linguistics}, pages 3346--3359, Santa Fe, New Mexico, USA. Association for Computational Linguistics.

\bibitem[{Thorne and Vlachos(2021)}]{thorne-vlachos-2021-evidence}
James Thorne and Andreas Vlachos. 2021.
\newblock \href {https://doi.org/10.18653/v1/2021.acl-long.256} {Evidence-based factual error correction}.
\newblock In \emph{Proceedings of the 59th Annual Meeting of the Association for Computational Linguistics and the 11th International Joint Conference on Natural Language Processing (Volume 1: Long Papers)}, pages 3298--3309, Online. Association for Computational Linguistics.

\bibitem[{Thorne et~al.(2018)Thorne, Vlachos, Christodoulopoulos, and Mittal}]{thorne-etal-2018-fever}
James Thorne, Andreas Vlachos, Christos Christodoulopoulos, and Arpit Mittal. 2018.
\newblock \href {https://doi.org/10.18653/v1/N18-1074} {{FEVER}: a large-scale dataset for fact extraction and {VER}ification}.
\newblock In \emph{Proceedings of the 2018 Conference of the North {A}merican Chapter of the Association for Computational Linguistics: Human Language Technologies, Volume 1 (Long Papers)}, pages 809--819, New Orleans, Louisiana. Association for Computational Linguistics.

\bibitem[{Vallayil et~al.(2023)Vallayil, Nand, Yan, and Allende-Cid}]{vallayil2023explainability}
Manju Vallayil, Parma Nand, Wei~Qi Yan, and H{\'e}ctor Allende-Cid. 2023.
\newblock Explainability of automated fact verification systems: A comprehensive review.
\newblock \emph{Applied Sciences}, 13(23):12608.

\bibitem[{Vaswani et~al.(2017)Vaswani, Shazeer, Parmar, Uszkoreit, Jones, Gomez, Kaiser, and Polosukhin}]{vaswani2017attention}
Ashish Vaswani, Noam Shazeer, Niki Parmar, Jakob Uszkoreit, Llion Jones, Aidan~N Gomez, {\L}ukasz Kaiser, and Illia Polosukhin. 2017.
\newblock Attention is all you need.
\newblock \emph{Advances in neural information processing systems}, 30.

\bibitem[{Vlachos and Riedel(2015)}]{vlachos-riedel-2015-identification}
Andreas Vlachos and Sebastian Riedel. 2015.
\newblock \href {https://doi.org/10.18653/v1/D15-1312} {Identification and verification of simple claims about statistical properties}.
\newblock In \emph{Proceedings of the 2015 Conference on Empirical Methods in Natural Language Processing}, pages 2596--2601, Lisbon, Portugal. Association for Computational Linguistics.

\bibitem[{Vladika and Matthes(2023)}]{vladika-matthes-2023-scientific}
Juraj Vladika and Florian Matthes. 2023.
\newblock \href {https://doi.org/10.18653/v1/2023.findings-acl.387} {Scientific fact-checking: A survey of resources and approaches}.
\newblock In \emph{Findings of the Association for Computational Linguistics: ACL 2023}, pages 6215--6230, Toronto, Canada. Association for Computational Linguistics.

\bibitem[{Wadden et~al.(2020)Wadden, Lin, Lo, Wang, van Zuylen, Cohan, and Hajishirzi}]{wadden-etal-2020-fact}
David Wadden, Shanchuan Lin, Kyle Lo, Lucy~Lu Wang, Madeleine van Zuylen, Arman Cohan, and Hannaneh Hajishirzi. 2020.
\newblock \href {https://doi.org/10.18653/v1/2020.emnlp-main.609} {Fact or fiction: Verifying scientific claims}.
\newblock In \emph{Proceedings of the 2020 Conference on Empirical Methods in Natural Language Processing (EMNLP)}, pages 7534--7550, Online. Association for Computational Linguistics.

\bibitem[{Wang and Shu(2023)}]{wang2023explainable}
Haoran Wang and Kai Shu. 2023.
\newblock Explainable claim verification via knowledge-grounded reasoning with large language models.
\newblock \emph{arXiv preprint arXiv:2310.05253}.

\bibitem[{Wang(2017)}]{wang-2017-liar}
William~Yang Wang. 2017.
\newblock \href {https://doi.org/10.18653/v1/P17-2067} {{``}liar, liar pants on fire{''}: A new benchmark dataset for fake news detection}.
\newblock In \emph{Proceedings of the 55th Annual Meeting of the Association for Computational Linguistics (Volume 2: Short Papers)}, pages 422--426, Vancouver, Canada. Association for Computational Linguistics.

\bibitem[{Wang et~al.(2023)Wang, Reddy, Mujahid, Arora, Rubashevskii, Geng, Afzal, Pan, Borenstein, Pillai et~al.}]{wang2023factcheck}
Yuxia Wang, Revanth~Gangi Reddy, Zain~Muhammad Mujahid, Arnav Arora, Aleksandr Rubashevskii, Jiahui Geng, Osama~Mohammed Afzal, Liangming Pan, Nadav Borenstein, Aditya Pillai, et~al. 2023.
\newblock Factcheck-gpt: End-to-end fine-grained document-level fact-checking and correction of llm output.
\newblock \emph{arXiv preprint arXiv:2311.09000}.

\bibitem[{Wei et~al.(2022)Wei, Wang, Schuurmans, Bosma, ichter, Xia, Chi, Le, and Zhou}]{NEURIPS2022_9d560961}
Jason Wei, Xuezhi Wang, Dale Schuurmans, Maarten Bosma, brian ichter, Fei Xia, Ed~Chi, Quoc~V Le, and Denny Zhou. 2022.
\newblock \href {https://proceedings.neurips.cc/paper_files/paper/2022/file/9d5609613524ecf4f15af0f7b31abca4-Paper-Conference.pdf} {Chain-of-thought prompting elicits reasoning in large language models}.
\newblock In \emph{Advances in Neural Information Processing Systems}, volume~35, pages 24824--24837. Curran Associates, Inc.

\bibitem[{Wu et~al.(2020)Wu, Rao, Zhao, Liang, and Nazir}]{wu-etal-2020-dtca}
Lianwei Wu, Yuan Rao, Yongqiang Zhao, Hao Liang, and Ambreen Nazir. 2020.
\newblock \href {https://doi.org/10.18653/v1/2020.acl-main.97} {{DTCA}: Decision tree-based co-attention networks for explainable claim verification}.
\newblock In \emph{Proceedings of the 58th Annual Meeting of the Association for Computational Linguistics}, pages 1024--1035, Online. Association for Computational Linguistics.

\bibitem[{Wu et~al.(2019)Wu, AbdAlmageed, and Natarajan}]{wu2019mantra}
Yue Wu, Wael AbdAlmageed, and Premkumar Natarajan. 2019.
\newblock Mantra-net: Manipulation tracing network for detection and localization of image forgeries with anomalous features.
\newblock In \emph{Proceedings of the IEEE/CVF Conference on Computer Vision and Pattern Recognition}, pages 9543--9552.

\bibitem[{Xu et~al.(2023)Xu, Liu, Wu, and Wang}]{xu-etal-2023-counterfactual}
Weizhi Xu, Qiang Liu, Shu Wu, and Liang Wang. 2023.
\newblock \href {https://doi.org/10.18653/v1/2023.acl-long.374} {Counterfactual debiasing for fact verification}.
\newblock In \emph{Proceedings of the 61st Annual Meeting of the Association for Computational Linguistics (Volume 1: Long Papers)}, pages 6777--6789, Toronto, Canada. Association for Computational Linguistics.

\bibitem[{Yang et~al.(2019)Yang, Pentyala, Mohseni, Du, Yuan, Linder, Ragan, Ji, and Hu}]{yang2019xfake}
Fan Yang, Shiva~K Pentyala, Sina Mohseni, Mengnan Du, Hao Yuan, Rhema Linder, Eric~D Ragan, Shuiwang Ji, and Xia Hu. 2019.
\newblock Xfake: Explainable fake news detector with visualizations.
\newblock In \emph{The world wide web conference}, pages 3600--3604.

\bibitem[{Yao et~al.(2023)Yao, Shah, Sun, Cho, and Huang}]{yao2023end}
Barry~Menglong Yao, Aditya Shah, Lichao Sun, Jin-Hee Cho, and Lifu Huang. 2023.
\newblock End-to-end multimodal fact-checking and explanation generation: A challenging dataset and models.
\newblock In \emph{Proceedings of the 46th International ACM SIGIR Conference on Research and Development in Information Retrieval}, pages 2733--2743.

\bibitem[{Zhang et~al.(2023{\natexlab{a}})Zhang, Liu, Zhang, Sun, Xie, and Zha}]{zhang2023ecenet}
Fanrui Zhang, Jiawei Liu, Qiang Zhang, Esther Sun, Jingyi Xie, and Zheng-Jun Zha. 2023{\natexlab{a}}.
\newblock Ecenet: Explainable and context-enhanced network for muti-modal fact verification.
\newblock In \emph{Proceedings of the 31st ACM International Conference on Multimedia}, pages 1231--1240.

\bibitem[{Zhang et~al.(2023{\natexlab{b}})Zhang, Li, Cui, Cai, Liu, Fu, Huang, Zhao, Zhang, Chen et~al.}]{zhang2023siren}
Yue Zhang, Yafu Li, Leyang Cui, Deng Cai, Lemao Liu, Tingchen Fu, Xinting Huang, Enbo Zhao, Yu~Zhang, Yulong Chen, et~al. 2023{\natexlab{b}}.
\newblock Siren's song in the ai ocean: A survey on hallucination in large language models.
\newblock \emph{arXiv preprint arXiv:2309.01219}.

\bibitem[{Zhu et~al.(2023)Zhu, Si, Zhao, Zhu, Zhou, and He}]{zhu2023explain}
Yingjie Zhu, Jiasheng Si, Yibo Zhao, Haiyang Zhu, Deyu Zhou, and Yulan He. 2023.
\newblock \href {http://arxiv.org/abs/2310.14508} {Explain, edit, generate: Rationale-sensitive counterfactual data augmentation for multi-hop fact verification}.

\end{thebibliography}
\bibliographystyle{acl_natbib}

\appendix
\section{Methodology for Literature Compilation}
\label{const}
This appendix outlines the methodology employed to compile the content of this survey. It details the search strategy and selection criteria used to curate the foundational content for this survey paper.
\begin{enumerate}
    \item \textbf{Search Strategy}. Initially, we conducted a comprehensive search in the ACL Anthology, Google Scholar, and Google Search for related surveys. Within the ACL Anthology, we focused on venues such as EMNLP, ACL, and NAACL. The search involved using keywords like fact-checking, fact-checking survey, misinformation detection, explainable facts, and automatic fact-checking. 
    
    Furthermore, we gathered surveys related to the production of justifications in AFC. Our goal was to identify the earliest and most frequently cited papers in these surveys, which we considered as foundational or "pioneer" papers. Afterward, we tracked all papers that referenced these pioneer works up until the date of submission.

    \item \textbf{Selection Criteria}. We only selected papers that were directly relevant to the subject matter of justification production in AFC. The selection was based on a careful review of the abstract, introduction, conclusion, and limitations of each paper. Following the selection phase, 73 relevant papers were chosen to form the foundational content of this paper.
\end{enumerate}


\end{document}